\documentclass{article}

\usepackage[preprint]{spconf}
\copyrightnotice{\copyright\ IEEE 2024}
\toappear{To appear in {\it ICIP 2024}}

\usepackage{spconf,amsmath,graphicx}
\usepackage{amssymb}
\usepackage{booktabs}

\usepackage{multirow}
\usepackage{xcolor}
\usepackage{url}
\usepackage[nocompress]{cite}
\let\oldbibitem\bibitem
\renewcommand{\bibitem}{%
    \setlength{\itemsep}{0.15\baselineskip}%
    \oldbibitem
}

\newcommand{\etal}{\textit{et al}.}

\title{Statistics-aware Audio-visual DeepFake Detector}
\name{Marcella Astrid$^1$ \qquad Enjie Ghorbel$^{1,2}$ \qquad Djamila Aouada$^1$ \thanks{This work was supported by the Luxembourg National Research Fund (FNR) under the project BRIDGES2021/IS/16353350/FaKeDeTeR and POST Luxembourg.
}}
\address{$^1$Interdisciplinary Centre for Security, Reliability and Trust, University of Luxembourg, Luxembourg \\
$^2$Cristal Laboratory, National School of Computer Sciences, Manouba University, Tunisia}

\begin{document}
%
\maketitle
\begin{abstract}
In this paper, we propose an enhanced audio-visual deep detection method.  Recent methods in audio-visual deepfake detection mostly assess the synchronization between audio and visual features. Although they have shown promising results, they are based on the maximization/minimization of isolated feature distances without considering feature statistics.  Moreover, they rely on cumbersome deep learning architectures and are heavily dependent on empirically fixed hyperparameters. Herein, to overcome these limitations, we propose: (1)  a statistical feature loss to enhance the discrimination capability of the model,  instead of relying solely on feature distances; (2) using the waveform for describing the audio as a replacement of frequency-based representations; (3) a post-processing normalization of the fakeness score; (4) the use of shallower network for reducing the computational complexity. Experiments on the DFDC and FakeAVCeleb datasets  demonstrate the relevance of the proposed method.

\end{abstract}
\begin{keywords}
deepfake detector, multi-modal, audio-visual, distribution, similarity
\end{keywords}

\vspace{-3mm}
\section{Introduction}
\label{sec:intro}
\vspace{-3mm}

The growth of the audio-visual deepfake technology has recently raised major concerns. Deepfakes, which are person-centric manipulated media, are becoming increasingly realistic. As a result, there is a pressing need to address their potential misuse, such as spreading misinformation and identity theft~\cite{kaliyar2022understanding}, by introducing effective audio-visual deepfake detection solutions. 


\addtocounter{footnote}{-1}

\begin{figure}[]
\begin{center}
   \includegraphics[width=\linewidth]{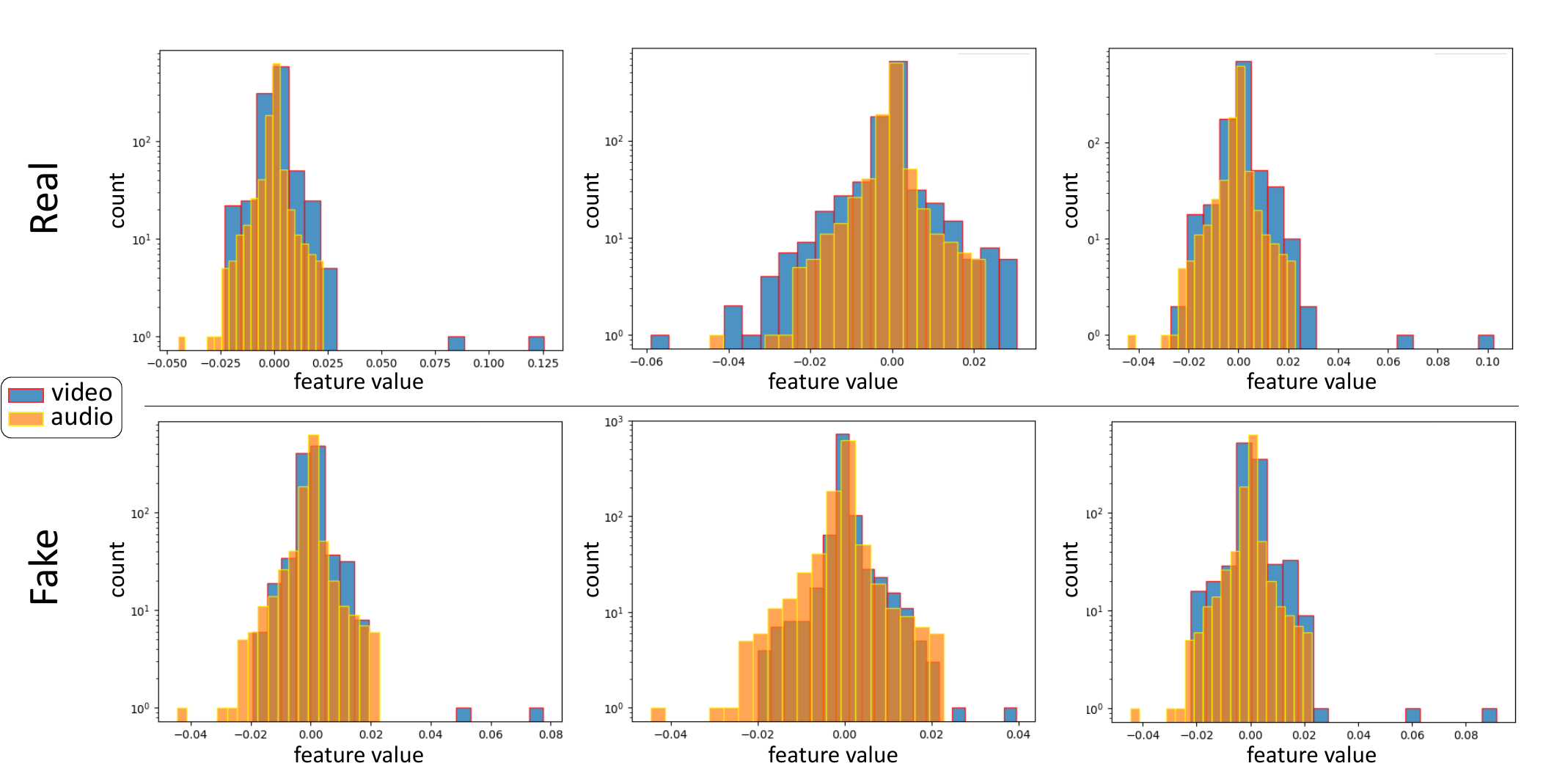}
\end{center}
\vspace{-5mm}
   \caption{Histograms of features extracted from three real and three fake samples using an enhanced version of MDS\protect\footnotemark. Each column represents the feature values in different range. }
   
\vspace{-4mm}
\label{fig:feat_vis_baseline_3sample}
\end{figure}

The most successful audio-visual deepfake detection methods involve the identification of inconsistencies between audio and visual features for isolating deepfakes~\cite{chugh2020not,gu2021deepfake,feng2023self}. They usually enforce a high distance between audio and visual features in the presence of fake data, while they aim to reduce this distance when dealing with real data. 

\addtocounter{footnote}{-1}

Despite being promising, these approaches might be improved from several perspectives: (1) existing methods mostly rely on isolated distances between audio and visual features extracted from one sample, ignoring feature statistics~\cite{chugh2020not,gu2021deepfake,mittal2020emotions}. As shown in Fig. \ref{fig:feat_vis_baseline_3sample}, histograms of visual and audio features extracted using our baseline\footnotemark ~from both real and fake test samples exhibit a similar pattern. In both cases, the audio and visual feature distributions are very similar, making the discrimination between real and fake data challenging; (2) existing methods are heavily dependent on empirically fixed hyperparameters. For instance, to model the audio modality, most deepfake detection methods employ a frequency representation \cite{chugh2020not,yang2023avoid,zhou2021joint}, potentially ignoring discriminative cues and necessitating a phase of hyper-parameter tuning, as highlighted in \cite{jung2019rawnet}. In addition, deepfakes are usually determined when the distance between the audio and visual features referred to as fakeness score exceeds a predefined threshold, similar to \cite{chugh2020not}; and (3)
Lastly, several audio-visual models incorporate very deep architectures, which can be computationally expensive \cite{chugh2020not,yang2023avoid}, making them unsuitable for numerous real-life applications.

\footnotetext{The baseline is the enhanced MDS without the statistics-aware loss discussed in Section \ref{subsec:baseline}. MDS \cite{chugh2020not} considers feature distance of whole face region frames and audio to detect deepfake.}

To address these challenges, we propose a novel shallow network called Statistics-aware Audio-visual Deepfake Detector (SADD) for audio-visual deepfake detection. In fact, recent works suggested that deepfake artifacts can be effectively modeled in low-level features extracted with shallower networks \cite{mejri2021leveraging,afchar2018mesonet}. To enhance the model's discrimination capability, instead of relying solely on a feature distance loss, we integrate a second loss that estimates the distance between first-order statistics of audio and visual feature distributions. This loss is then maximized for fake data, and minimized otherwise. Moreover, the raw waveform is employed as input to mitigate potential limitations associated with frequency-based representations. Finally, to avoid experimentally fixing a classification threshold, we suggest a fakeness score normalization. The experiments carried out on two well-known datasets, namely DFDC and FakeAVCeleb, show that our method necessitates a lower computational cost as compared to existing methods while maintaining competitive deepfake detection capabilities.

In summary, our contributions can be outlined as follows: (1) Taking the model from \cite{chugh2020not}, we enhance it by changing the audio input to waveform, reducing the network depth, and modifying the post-processing steps; (2) We identify an issue in such models, specifically the lack of distinguishable differences between real and fake distributions; (3) We introduce a novel loss function to enforce a desirable separation of audio-visual distributions for fake data and, conversely, promote their similarity for real data; (4) We conduct extensive evaluations of our model on the widely-used DFDC \cite{dolhansky2020deepfake} benchmark dataset, demonstrating superior performance compared to state-of-the-art (SoA) methods. Additionally, we observe an increase in generalization capabilities when applying our proposed component in a cross-dataset setting, utilizing the FakeAVCeleb dataset \cite{khalid2021fakeavceleb}.

\noindent \textbf{Organization of the paper:} Section \ref{sec:relatedworks} describes the related works. Section \ref{sec:methodology} introduces the proposed approach. Section \ref{sec:experiments} reports the experiments. Section \ref{sec:conclusion} concludes the paper.

\vspace{-3mm}
\section{Related works}
\label{sec:relatedworks}
\vspace{-2mm}


Several researchers have developed deepfake detectors for various modalities, such as images  \cite{chen2022self,afchar2018mesonet} and audios \cite{jung2022aasist,desplanques2020ecapa}. Nevertheless, these methods are mostly based on singular modalities, ignoring the complementary information incorporated in multi-modal data. In fact, leveraging multiple modalities is a promising way to obtain more robust deepfake detectors. Hence, numerous audio-visual deepfake detectors have been recently introduced to simultaneously consider audio and visual information extracted from a given video.

Existing audio-visual methods can be categorized into three classes, namely, identity-based, fusion-based, and inconsistency-based. Identity-based methods aim to detect manipulated videos of specific individuals. Agarwal \etal \cite{agarwal2023watch} explore the correlation between observed behavior and speech, enabling not only the detection of deepfakes but also the identification of doppelgangers (real persons with highly similar appearances). 
Cheng \etal \cite{cheng2023voice} attempt to ascertain if a voice matches a facial image to identify deepfakes. However, relying solely on images as visual cues may restrict detection to individuals within the dataset. 
Cai \etal \cite{cai2022you} define antonyms of real words as fake data. Nonetheless, the meaning of a sentence is heavily dependant on a person's personality. While detecting deepfakes of specific individuals has its specific applications, our interest lies in a more generic approach capable of robustly detecting deepfakes of unknown persons.

Alternatively, several fusion-based methods simply concatenate information from both audio and visual sources. Yang \etal \cite{yang2023avoid} combines audio and visual features using cross-attention after separately extracting audio and visual features using a transformer-based model. 
Zhou and Lim \cite{zhou2021joint} aggregate audio and visual multi-level features after applying cross-attention.
Lewis \etal \cite{lewis2020deepfake} combine multiple types of visual information (e.g., lip reading, color) and audio (e.g., spectrogram, phoneme) features using Long 
Short-Term Memory networks and a multi-layer perceptron. 
Ilyas \etal \cite{ilyas2023avfakenet} combines the decisions of the audio and visual models using a voting mechanism, which can be more sensitive to false positives, as a false positive from either the visual or audio model may result in an overall false positive prediction. 

Inconsistency-based methods explicitly focus on the synchronization between audio and visual elements to detect deepfakes. The assumption is that, in fake data, the audio fails to synchronize correctly with visual cues. Gu \etal \cite{gu2021deepfake} focus solely on the lips region as visual input, using a contrastive loss to train the (a)synchronization of visual and audio features. However, relying exclusively on the lips may lead to the omission of other crucial facial regions. Mittal \etal \cite{mittal2020emotions} specifically investigate the mismatch between emotions  that are visually perceived and those conveyed through audio/speech. Nevertheless, this approach relies on emotion recognition models. Feng \etal \cite{feng2023self} employ a synchronization model to predict whether a pair of visual and audio inputs is synchronized. 
This model utilizes pseudo-fake data generated by translating the real audio/visual information to the time domain, assuming that only translation mismatches exist in fake data.
Chugh \etal \cite{chugh2020not} minimizes the distance between audio and visual features extracted from real data while maximizing it in other cases. 
To enforce a minimum distance on real data while maximizing it in fake data, MDS utilizes contrastive learning. Contrastive learning has been applied in various domains, including face verification \cite{chopra2005learning} and lip reading \cite{chung2017out}. Compared to the aforementioned methods, MDS is straightforward, independent of other models, utilizes the entire face, and does not solely rely on translation synchronization assumptions. 
Nevertheless, as mentioned in Section 1, MDS only align pair of features independently, without considering their statistics which can lead to limited separation between audio and visual features extracted from fake data. As a solution, we introduce a statistics-aware loss to further enhance the separability between audio and visual features when encountering fake data.  

\vspace{-3mm}
\section{Methodology}
\label{sec:methodology}
\vspace{-2mm}

\begin{figure}[]
\begin{center}
   \includegraphics[width=\linewidth]{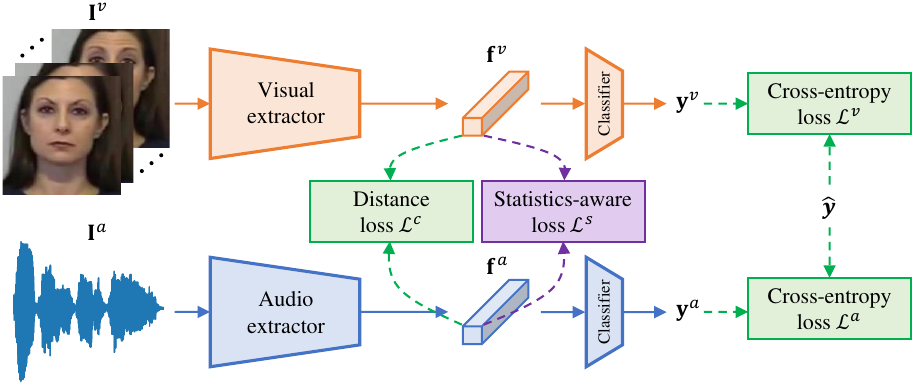}
\end{center}
    \vspace{-5mm}
   \caption{Our method consists of  visual and audio feature extractors that extract respectively features $\mathbf f^v$ and $\mathbf f^a$ from image sequences $\mathbf I^v$ and audio waveforms $\mathbf I^a$. Two separate classification layers are integrated on top of each extractor. The model is trained using a cross-entropy loss for each network, along with a feature distance loss. To enhance feature discrimination between real and fake data, we introduce an additional feature statistics-aware loss.}
   
\label{fig:method}
\end{figure}

We propose a multi-modal deepfake detector which exploits the inconsistencies between audio and visual features for detecting deepfakes. 
Fig. \ref{fig:method} gives an overview of the proposed approach. In particular, the architecture proposed by Chugh \etal \cite{chugh2020not} (Section \ref{subsec:baseline}) is enhanced. Moreover, we employ an additional loss to quantify the first-order statistics of real and fake data, hence improving the discriminative power of audio and visual features (Section \ref{subsec:distribution_loss}).

\subsection{Enhanced Modality
Dissonance Score}
\label{subsec:baseline}

To better leverage the inconsistencies between audio and visual cues for deepfake detection, we seek a model that is simple, free from assumptions, independent of other models, and capable of utilizing entire facial regions without discarding potentially important artifacts. Such a model offers flexibility for improvements. We find the Modality
Dissonance Score (MDS)~\cite{chugh2020not} approach to fit our requirements and propose some modifications for enhancing it.
To reduce the computational cost, we first suggest the use of a shallow version of MDS by reducing its depth. As mentioned in Section \ref{sec:intro}, several works have shown that low-level representations are sufficient for effectively modeling deepfake artifacts \cite{chugh2020not}, implying the relevance of shallower networks. Additionally, to reduce the need for hyperparameter tuning related to audio input conversion, we use the waveform  representation as audio input and introduce a refinement strategy of the fakeness score during inference. More details about these steps are provided in next subsections.

\vspace{-1mm}
\subsubsection{Architecture}
\vspace{-1mm}
We extract features separately for audio input $ \mathbf I^a$ and visual input $ \mathbf I^v$ using the audio and the visual feature extractor denoted as $\mathfrak{A}$ and $\mathfrak{V}$, respectively, as follows,
\begin{equation}
    \mathbf f^a = \mathfrak{A}( \mathbf  I^a) \quad\text{and}\quad   \mathbf  f^v = \mathfrak{V}(  \mathbf I^v) \text{,}  
\end{equation}
where $f^a$ and $f^v$ are vectors of the same size.

In \cite{chugh2020not}, $\mathbf I^a$ is represented by a Mel spectogram. However, converting the waveform to a Mel spectogram may introduce limitations from the conversion process, e.g., the necessity of manually fixing hyperparameters such as the window size, the shift, length, and the feature dimension which can be sub-optimal \cite{jung2019rawnet}. As an alternative, we represent the features $\mathbf I^a$ using the raw waveform. Consequently, $\mathfrak{A}$ is 
adapted to comply with this kind of input. Specifically, we use 1D convolutions instead of the 2D convolution layers employed in \cite{chugh2020not}. Furthermore, we posit that deepfake artifacts are present in shallow features, hence opting for a reduced number of convolution layers. Specifically, the size of the visual extractor is reduced from five to three convolution blocks. Meanwhile, the audio extractors reduced from four to two convolution blocks. However, extracting features from shallower network can lead to larger feature maps, which can increase the computational cost in the fully connected layers. Therefore, we retain only the max pooling from the later blocks of the deep network to downsample the extracted feature maps.

For real/fake classification, two separated fully connected layers $\mathfrak{C}^a$ and $\mathfrak{C}^v$ are considered on top of the audio and visual extractors as detailed below, respectively,
\begin{equation}
    \mathbf y^a = \mathfrak{C}^a(   \mathbf  f^a) \quad\text{and}\quad     \mathbf y^v = \mathfrak{C}^v(   \mathbf f^v) \text{,}
\end{equation}
where $  \mathbf y^a$ and $   \mathbf  y^v$ represent the class predictions (i.e., real/fake) for the audio and visual branches, respectively.

\vspace{-1mm}
\subsubsection{Training}
\vspace{-1mm}

To train the proposed network, the three losses introduced in \cite{chugh2020not}, namely, the audio cross-entropy loss $\mathcal L^a$, the visual cross-entropy loss $\mathcal  L^v$, and the contrastive-based distance loss $\mathcal  L^c$ are used. Moreover, we propose an additional based loss taking into account the feature statistics denoted as  $\mathcal L^{s}$. More details about the calculation of $\mathcal L^{s}$ is given in Section \ref{subsec:distribution_loss}. The total loss $\mathcal L$ is then computed as follows,
\begin{equation}
    \mathcal L =  \mathcal L^v +\mathcal  L^a + \mathcal L^c + \alpha \mathcal L^s  \text{,}
\label{eq:final_loss}
\end{equation}
where $\mathcal L^a$ and $\mathcal L^v$ represents the cross-entropy loss between $\mathbf y^a$ and $\hat{\mathbf  y}$, and $\mathbf y^v$ and $\hat{\mathbf  y}$, respectively. The variable $\alpha$ is a weighting hyperparameter. Despite the presence of noisy labels (i.e., $\hat{\mathbf  y}$ is ``fake'' in audio-visual may mean only-audio, only-visual, or both modalities being fake), Chugh \etal \cite{chugh2020not} experimentally demonstrates the importance of these cross-entropy losses.

The contrastive loss enforces consistent audio-visual features for real data and inconsistency for fake data as follows,
\begin{equation}
    \mathcal L^c= 
\begin{cases}
    (d^{\mathbf f})^2 & \text{if } \mathbf y \text{ is real}\text{,}\\
    (\max(m-d^{\mathbf f}, 0))^2 & \text{if } \mathbf y \text{ is fake}\text{,}\\
\end{cases}
\end{equation}
where $m=0.99$ is the margin, as defined in \cite{chugh2020not}, and $d^{\mathbf f}$ is the squared $L_2$ distance between $\mathbf f^v$ and $\mathbf f^a$.






\vspace{-1mm}
\subsubsection{Inference}
\label{subsubsec:inference}
\vspace{-1mm}
The fakeness score denoted as $\mu^d$ of a test video is determined based on the mean distance calculated for all subsequences contained in the test video, as expressed by the following formula,
\begin{equation}
    \mu^d = \frac{1}{n_s} \sum_{i=1}^{n_s} d^{\mathbf f}_i \text{,}
\end{equation}
where $n_s$ represents the number of subsequences in a video and $d^{\mathbf f}_i $ is $d^{\mathbf f}$ of the $i$-th sequence.


In order to constrain the fakeness score to vary between 0 to 1 for a more intuitive prediction, Chugh \etal \cite{chugh2020not} implement a post-processing to derive the final $s$ score as follows,
\begin{equation}
    s= 
\begin{cases}
    1 \text{ (real) } & \text{if } \mu^d < \tau \text{,}\\
    0 \text{ (fake) } & \text{if } \mu^d \geq \tau\text{,}\\
\end{cases}
\end{equation}
where $\tau$ is selected by finding the threshold with the highest performance in the training set.

However, exhaustively searching for the best threshold value may hide the full potential of the model. Instead, we min-max normalize the score using the minimum and maximum values of $\mu^d$ from the training data. Thus, for a given test video, the score is calculated as,
\begin{equation}
    s = \frac{\mu^d - \mu^d_{min\_train}}{\mu^d_{max\_train}- \mu^d_{min\_train}} \text{,}
\end{equation}
where $\mu^d_{min\_train}$ represents the minimum of $\mu^d$ across the training set, and $\mu^d_{max\_train}$ is the maximum. The score is then clipped to ensure it falls within the range of $0$ to $1$ as follows,
\begin{equation}
    s =  \min(1, \max(s, 0)) \text{.}
\end{equation}


\vspace{-2mm}
\subsection{Statistics-aware Loss}
\label{subsec:distribution_loss}
\vspace{-1mm}

Given the unclear distinction between the distributions of audio and visual features of  fake data as illustrated in Fig. \ref{fig:feat_vis_baseline_3sample}, we propose an additional loss $\mathcal L^s$ to enhance the discriminative power of  audio and visual features. 


In this work, we suggest to maximize the distance between first-order statistics (mean) of $\mathbf f^a$ and $\mathbf f^v$ denoted as $\mu^a$ and $\mu^v$, respectively, in the presence of fake data, while simultaneously minimizing it when dealing with real data. This is achieved using a contrastive loss defined as,
\begin{equation}
   \mathcal L^s= 
\begin{cases}
    (d^{\mu})^2 & \text{if } y \text{ is real}\text{,}\\
    \max(m^\mu-(d^{\mu})^2, 0) & \text{if } y \text{ is fake}\text{,}\\
\end{cases}
\label{eq:mean_separation_loss}
\end{equation}
where $d^\mu$ is the $L_2$ distance between $\mu^a$ and $\mu^v$. In this way, $d^\mu$ should be low when encountering a real video and high otherwise.

Regarding the margin $m^\mu$, while it could be set as a constant hyperparameter, we propose an alternative adaptive approach. Specifically, we suggest using the standard deviation of $\mathbf f^a$ and $\mathbf f^v$  as the margin,
\begin{equation}
    m^\mu = \sigma^v + \sigma^a \text{,}
\label{eq:margin_with_std}
\end{equation}
where $\sigma^a$ and $\sigma^v$ represent the standard deviations of $\mathbf f^a$ and $\mathbf f^v$, respectively. Leveraging the standard deviation information instead of using a predefined hyperparameter serves two purposes. Firstly, it helps reduce the number of predefined hyperparameters. Secondly, it allows for a dynamic determination of how far apart two mean values need to be separated based on the standard deviation. In other words, when separating two distributions, those with a smaller standard deviation may not require their means to be separated as far as distributions with a larger standard deviation.

\vspace{-3mm}
\section{Experiments}
\label{sec:experiments}
\vspace{-2mm}

\subsection{Dataset}
\vspace{-1mm}

\noindent \textbf{DFDC \cite{dolhansky2020deepfake}.} We employ the DeepFake Detection Challenge (DFDC) dataset for both training and testing purposes, following the protocol described in \cite{chugh2020not,mittal2020emotions}, which employs a subset of $18,000$ videos. This subset is divided into train and test sets with an $85:15$ ratio. 
Additionally, for accelerated initial experiments, we also create a smaller set of training data consisting of $4,930$ videos.
Unless specified otherwise, we utilize the larger set.

\noindent \textbf{FakeAVCeleb \cite{khalid2021fakeavceleb}.} We exclusively use FakeAVCeleb for testing our DFDC-trained models to assess their generalization capabilities to a new dataset. We adhere to the test setup outlined in \cite{khalid2021evaluation}, where a test set comprises $70$ fake videos and $70$ real videos.

\vspace{-2mm}
\subsection{Experiment Setup}
\vspace{-1mm}
We generally follow the setup proposed in the official code of \cite{chugh2020not}\footnote{\url{https://github.com/abhinavdhall/deepfake/tree/main/ACM_MM_2020}}, except converting audio to a Mel spectogram.
Approximately $1$ seconds of audio input ($\mathbf I^a$) is represented as a waveform with a size of $T^a \times C^a = 48000 \times 1$, where $T^a$ and $C^a$ denote the time and channel size, respectively. The video input sequence ($\mathbf I^v$) is of size $T^v \times C^v \times H^v \times W^v = 30 \times 3 \times 224 \times 224$, where $T^v$, $C^v$, $H^v$, and $W^v$ are the time, channel, height, and width sizes. This video input is extracted concurrently with $\mathbf I^a$ but at a different sampling rate ($30$ instead of $48000$). 
The model is trained for $50$ epochs with the Adam optimizer \cite{kingma2014adam}, a learning rate of $10^{-3}$, a batch size of $8$, and a weight decay of $10^{-5}$. The model with the lowest total loss throughout the training epoch is saved for evaluation. Unless specified otherwise, we set the weighting hyperparameter $\alpha$ in Eq. \eqref{eq:final_loss} to $1$. We report the widely-used Area Under the Curve (AUC) metric to measure the performance of the proposed model \cite{chugh2020not,mittal2020emotions}.

\vspace{-2mm}
\subsection{Ablation Study}
\label{subsec:ablation_studies}
\vspace{-1mm}


To enhance the model proposed in \cite{chugh2020not}, we introduce three modifications: 1) post-processing by normalizing the fakeness score based on the score statistics in the training set (Section~\ref{subsubsec:inference}); 2) changing audio input type from frequency-based to waveform; and (3) reducing the network size. The significance of normalization with the training set in post-processing is evident when comparing Table \ref{tab:baseline_comparisons}(a) and Table \ref{tab:baseline_comparisons}(b). Transitioning from Table \ref{tab:baseline_comparisons}(b) to Table \ref{tab:baseline_comparisons}(c), we change the audio input type from Mel spectrogram to waveform, resulting in an improved performance. Reducing the network size in Table \ref{tab:baseline_comparisons}(d) decreases the number of parameters by more than half while maintaining comparable performance, showcasing that a shallower network is sufficient for deepfake detection. The model in Table \ref{tab:baseline_comparisons}(d) is defined as our baseline from here forth. Incorporating the statistics-aware loss in Table \ref{tab:baseline_comparisons}(e) further enhances performance without introducing additional parameters. These results underscore the importance of each proposed component.

\begin{table}[]
\centering
\caption{Ablation study: (a) original model of \cite{chugh2020not}, (b) with normalization in post-processing, (c) with audio format change to waveform, (d) with shallower network (baseline), and (e) with added distribution loss. ``Size'', ``Audio'', ``Norm.'', ``$\mathcal L^s$'', ``AUC'', ``\#Param'' represent network size, audio input type (Mel: Mel-spectogram; Wave.: Waveform), normalization in post-processing, statistics-aware loss, AUC on DFDC, and number of model parameters in millions, respectively.}
\resizebox{\linewidth}{!}{
\begin{tabular}{|c|l|l|c|c||c|c|c|}
\hline
 & Size & Audio & Norm. & $\mathcal L^s$ & AUC & \#Param \\ \hline
 (a) & Deep                          & Mel   &   &          & 89.87\%                       & 122.78\\
 (b) & Deep                          & Mel   & \checkmark  &          & 96.34\%                 & 122.78\\
 (c) & Deep                          & Wave.   & \checkmark  &          & 97.13\%             & 263.67 \\
 (d) & Shallow                       & Wave.     & \checkmark   &                 & 96.55\%         & 107.10\\ \hline
 (e) & Shallow                       & Wave.      & \checkmark  & \checkmark                            &    96.69\%    &  107.10   \\ \hline
\end{tabular}
}
\vspace{-3mm}
\label{tab:baseline_comparisons}
\end{table}

\vspace{-2mm}
\subsection{Comparisons with SoA}
\label{subsec:comparisonswithsoa}
\vspace{-1mm}

Table \ref{tab:soa} compares  our approach with SoA methods in terms of AUC on DFDC. When compared to uni-modal methods, either only-visual (V) or only-audio (A), multi-modal methods generally exhibit better performance. In the comparison among audio-visual models, our method demonstrates superiority over other models, including the MDS \cite{chugh2020not} baseline. We also outperform methods using emotion-based inconsistencies \cite{mittal2020emotions}. Furthermore, our performance surpasses identity-based models, such as BA-TFD \cite{cai2022you} and VFD \cite{cheng2023voice}. Additionally, we outperform fusion-based models, such as AVoiD-DF \cite{yang2023avoid} and AVFakeNet \cite{ilyas2023avfakenet}.

\begin{table}[]
\centering
\caption{Comparison of  the proposed model with SoA methods in terms of AUC on the DFDC dataset using only-visual (V), only-audio (A), and audio-visual (AV) modalities.}
\resizebox{\linewidth}{!}{
\begin{tabular}{lcc||l|c|c|}
\hline
\multicolumn{1}{|l|}{Method}                                                  & \multicolumn{1}{c|}{Modality} & AUC                   & Method                                               & Modality & AUC     \\ \hline
\multicolumn{1}{|l|}{Multi-Attention \cite{zhao2021multi}}   & \multicolumn{1}{c|}{V}        & 84.8\%                & MDS \cite{chugh2020not}             & AV       & 90.66\% \\ 
\multicolumn{1}{|l|}{SLADD \cite{chen2022self}}             & \multicolumn{1}{c|}{V}        & 75.2\%                & Emotion \cite{mittal2020emotions}   & AV       & 84.4\%  \\ 
\multicolumn{1}{|l|}{Meso4 \cite{afchar2018mesonet}}         & \multicolumn{1}{c|}{V}        & 75.3\%                & BA-TFD \cite{cai2022you}            & AV       & 84.6\%  \\ \cline{1-3}
\multicolumn{1}{|l|}{AASIST \cite{jung2022aasist}}           & \multicolumn{1}{c|}{A}        & 68.4\%                & AVFakeNet \cite{ilyas2023avfakenet} & AV       & 86.2\%  \\ 
\multicolumn{1}{|l|}{ECAPA-TDNN \cite{desplanques2020ecapa}} & \multicolumn{1}{c|}{A}        & 69.8\%                & VFD \cite{cheng2023voice}           & AV       & 85.13\% \\ 
\multicolumn{1}{|l|}{RawNet \cite{jung2019rawnet}}           & \multicolumn{1}{c|}{A}        & 56.2\%                & AvoiD-DF \cite{yang2023avoid}       & AV       & 94.8\%  \\ \cline{1-3}
                                                                              & \multicolumn{1}{l}{}          & \multicolumn{1}{l|}{} & SADD (Ours)                                          & AV       & \textbf{96.69\%} \\ \cline{4-6} 
\end{tabular}
}
\vspace{-3mm}
\label{tab:soa}
\end{table}

\vspace{-1mm}
\subsection{Additional Discussions}

\vspace{-1mm}
\subsubsection{How does $\alpha$ affect the model?}
\vspace{-1mm}



We introduce a weighting hyperparameter $\alpha$ in Eq. \eqref{eq:final_loss} to adjust the contribution of the statistics-aware loss within the overall loss. We report the obtained results on a smaller training set, when varying the value of $\alpha$ . As shown in Table \ref{tab:hyperparam}, for different values of $\alpha$ , the model achieves better performance compared to the baseline. However, setting $\alpha$ to relatively large value results in a significant drop in AUC.

\begin{figure*}[]
\begin{center}
   \includegraphics[width=0.95\linewidth]{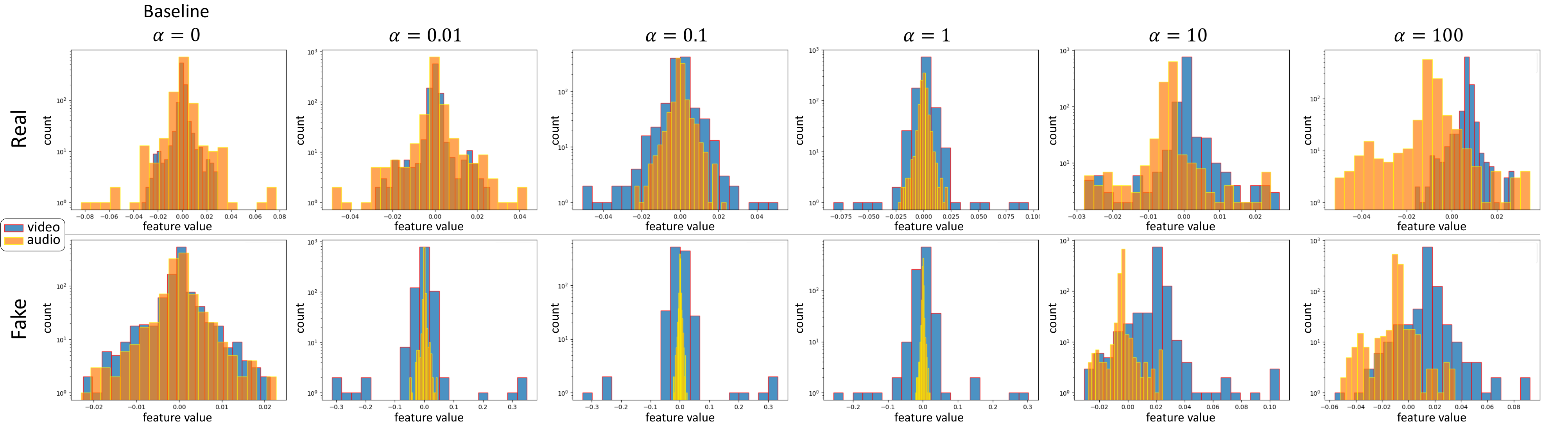}
\end{center}
\vspace{-7mm}
   \caption{
   Feature distribution histograms of real and fake data with different values of $\alpha$. Adding the statistics-aware loss ($\alpha > 0$) results in distinguishable distribution characteristics between real and fake data. 
   }
\vspace{-5mm}
\label{fig:feat_vis_msl_hyperparam}
\end{figure*}

To further analyze the behavior of the model with different $\alpha$ values, we visualize in Fig.~\ref{fig:feat_vis_msl_hyperparam} the distribution of audio and visual features from a real and a fake test sample. Interestingly, for most $\alpha$ values, instead of having separate mean values between audio and visual features for fake samples, the distribution exhibits different standard deviations. This can be attributed to the margin $m^{\mu}$ in Eq. \eqref{eq:mean_separation_loss}, which is defined using the standard deviation (Eq. \eqref{eq:margin_with_std}). Consequently, the loss in Eq. \eqref{eq:mean_separation_loss} can decrease if the total standard deviation becomes smaller. Hence, a smaller standard deviation of the audio distribution can lead to a smaller $L^s$. Nevertheless, the observed distinct characteristics between real and fake data may contribute to the overall improvement of the model. Instead of observing a difference in terms of standard deviation, a sseperation strats to appear between the mean values of audio and visual features, with larger $\alpha$ values ($\alpha \geq 10$). However, a high value of $\alpha$ (i.e., $\alpha = 100$) can also impact the distribution of real data, making the real and fake data distributions indistinguishable again.

\begin{table}[]
\centering
\caption{Evaluation on DFDC across a wide range of $\alpha$ values. Each model is trained on the smaller training set. In general, our model is robust to a huge range of $\alpha$ values, achieving better performance compared to the baseline.}
\begin{tabular}{|c|c||c|c|}
\hline
$\alpha$ & AUC   & $\alpha$ & AUC   \\ \hline
0 (baseline)    & 86.52\% & 1     & \textbf{88.55\%} \\
0.01  & 87.33\% & 5     & 86.64\% \\
0.05  & 88.14\% & 10    & 86.92\% \\
0.1   & 86.57\% & 50    & 86.59\% \\
0.5   & 87.03\% & 100   & 83.51\% \\ \hline
\end{tabular}
\vspace{-3mm}
\label{tab:hyperparam}
\end{table}

\vspace{-1mm}
\subsubsection{Impact of the statistics-aware loss in limited training data}
\vspace{-1mm}
We train the models detailed in Table \ref{tab:hyperparam} using the smaller training dataset. The model incorporating the  statistics-aware loss achieves an improvement of $2\%$ as compared to the baseline, highlighting the relevance of this loss in a limited training data environment.


\vspace{-1mm}
\subsubsection{Cross-dataset generalization}
\vspace{-1mm}

To assess the generalisation capabilities of the proposed method, we test our model (trained on the dataset DFDC) on the FakeAVCeleb dataset. 
The models reported in Table \ref{tab:baseline_comparisons}(a)-(e) achieve AUC scores of $48.57\%$, $56.92\%$, $59.87\%$, $58.82\%$, and $61.39\%$, respectively. The noticable enhancement in performance suggests improved generalization capabilities resulting from the utilization of waveform and the incorporation of statistics-aware loss. 
However, the proposed model still lags behind SoA, such as AvoiD-DF \cite{yang2023avoid} with $82.8\%$ AUC, prompting the need for further investigations and refinements. One potential avenue for improvement could be exploring the benefits of a transformer-based model, as utilized in AvoiD-DF.

\vspace{-2mm}
\subsubsection{Is the behavior induced by the statistics-aware loss consistent?}
\begin{figure}[]
\begin{center}
   \includegraphics[width=\linewidth]{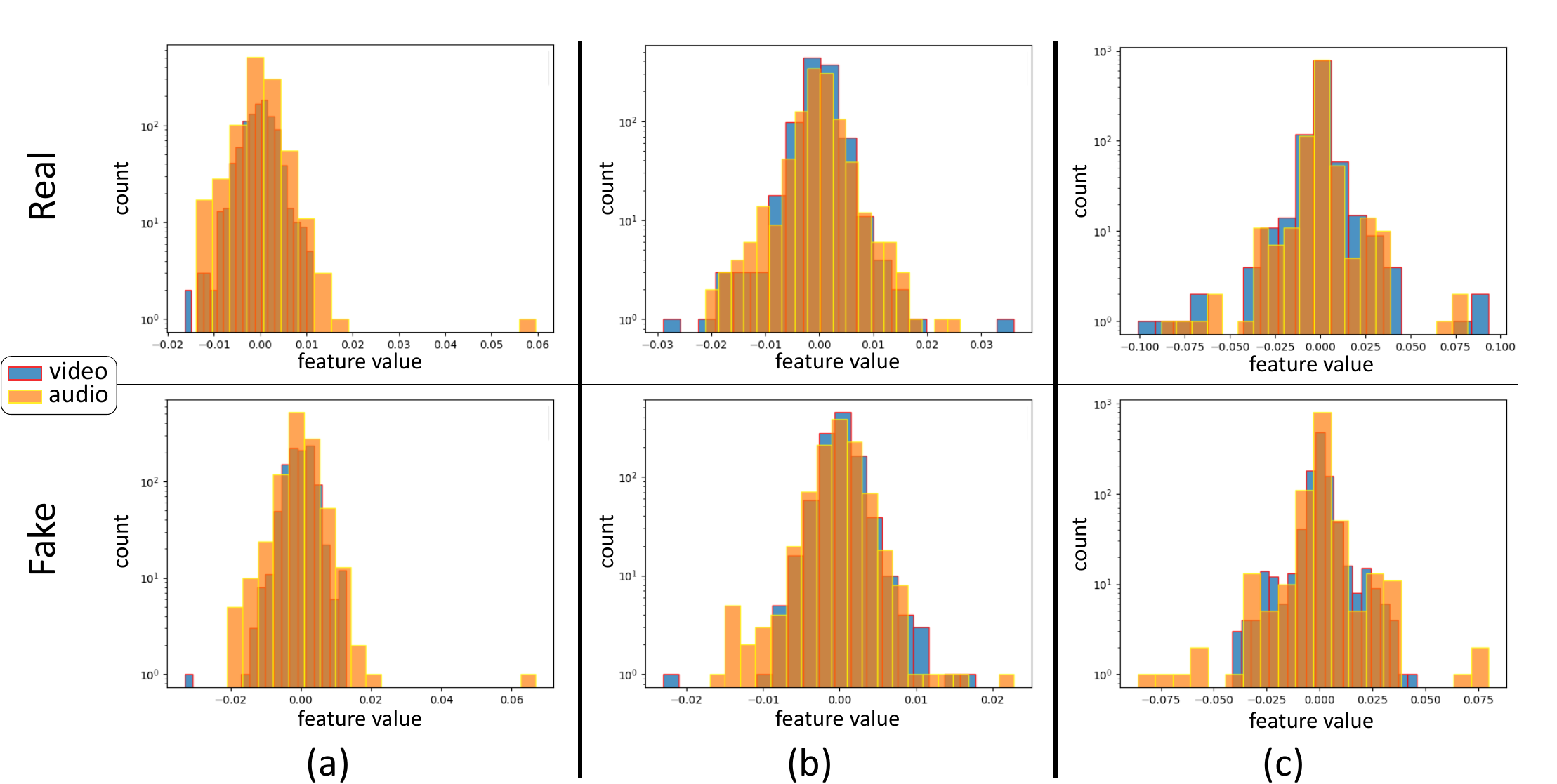}
\end{center}
\vspace{-6mm}
   \caption{Phenomena observed in Fig. \ref{fig:feat_vis_baseline_3sample} are also evident in other models without statistics-aware loss: (a) Deep network with mel-spectrogram audio input (Table \ref{tab:baseline_comparisons}(b)); (b) Deep network with waveform audio input (Table \ref{tab:baseline_comparisons}(c)); (c) Shallow network with waveform audio input (Table \ref{tab:baseline_comparisons}(d)) trained with the smaller set.}
\vspace{-5mm}
\label{fig:feat_vis_othermodels}
\end{figure}

\vspace{-1mm}
To further emphasize the importance of the statistics-aware loss, we visualize in addition to Fig. \ref{fig:feat_vis_baseline_3sample}, the feature distributions extracted from models with different settings. Fig. \ref{fig:feat_vis_othermodels}(a) and (b) correspond to the models described in Table \ref{tab:baseline_comparisons}(b) and (c), respectively. Fig. \ref{fig:feat_vis_othermodels}(c) represents our baseline trained on a smaller set. As observed, a similar trend occurs in different models, indicating that the proposed statistics-aware loss has the potential to enhance their performance.

\vspace{-2mm}
\subsubsection{Consideration of KL divergence}
\vspace{-1mm}
One well-known metric for measuring the distance between two distributions is Kullback–Leibler (KL) divergence. However, utilizing KL divergence directly in our method requires backpropagation through feature histogram generation, which is not straightforward. For experimental purposes involving KL divergence, we treat each feature vector as a probability distribution by applying \textit{log\_softmax} $l(\cdot)$ to $\mathbf f^a$ and $\mathbf f^v$. Consequently, the KL distance loss is defined as,

\begin{equation}
    \mathcal L^s= \exp l(f^a) \log \frac{l(\mathbf f^a)}{l(\mathbf f^v)} \text{,}
\label{eq:kl_loss}
\end{equation}

For a fair comparison, similar to the proposed statistics-aware loss, we determine the optimal $\alpha$ hyperparameter value using a small training set and find $\alpha=0.05$. We then retrain the entire model with the full training set. However, the model trained using the KL divergence alone achieved an AUC of $95.91\%$ failing to surpass the baseline with an $96.55\%$ AUC. Alternative approaches, such as reinforcement learning to approximate the gradients, could be considered in future work.


\vspace{-3mm}
\section{Conclusion}
\label{sec:conclusion}
\vspace{-2mm}

In this paper, an enhanced version of MDS \cite{chugh2020not} has been introduced. MDS which aims at assessing whether the audio is synchronized with the visual content has registered promising results. Nevertheless, such an approach is subject to several drawbacks. Namely, it depends on numerous empirically fixed hyperparameters, does not consider the overall feature statistics and is based on a cumbersome architecture. To address that,  a statistics-aware loss  that enforces the separation of feature distributions for fake data while bringing them closer together for real data has been proposed. In addition, a shallower architecture relying on waveform as input has been introduced. Finally, an inference post-processing strategy has been followed. Evaluations on the DFDC and the FakeAVCeleb datasets, confirms the significance of the statistics-aware. In future work, more suitable and complete distribution losses will be considered.

\bibliographystyle{ieeetr}
\bibliography{refs}

\end{document}